\documentclass[10pt,twocolumn,letterpaper]{article}

\usepackage{cvpr}
\usepackage{times}
\usepackage{epsfig}
\usepackage{graphicx}
\usepackage{amsmath}
\usepackage{amssymb}
\usepackage{color}
\usepackage{multirow}
\usepackage{subcaption}
\usepackage{threeparttable}
\usepackage{authblk}

\newcommand*{\dif}{\mathop{}\!\mathrm{d}}

\usepackage[breaklinks=true,bookmarks=false]{hyperref}

\cvprfinalcopy 

\def\cvprPaperID{****} 

\begin{document}

\title{Generalized Focal Loss V2: Learning Reliable Localization Quality Estimation for Dense Object Detection} 

\author[]{Xiang Li$^{1,2}$, Wenhai Wang$^{3}$, Xiaolin Hu$^{4}$, Jun Li$^{1}$, Jinhui Tang$^{1}$, and Jian Yang$^{1}$\thanks{Corresponding author. Xiang Li, Jun Li and Jian Yang are from PCA Lab, Key Lab of Intelligent Perception and Systems for High-Dimensional Information of Ministry of Education, and Jiangsu Key Lab of Image and Video Understanding for Social Security, School of Computer Science and Engineering, Nanjing University of Science and Technology. \\ \\  $^{1}$Nanjing University of Science and Technology $^{2}$Momenta $^{3}$Nanjing University $^{4}$Tsinghua University \scriptsize \{xiang.li.implus, jinhuitang, csjyang\}@njust.edu.cn, wangwenhai362@163.com, xlhu@mail.tsinghua.edu.cn, junl.mldl@gmail.com}}

\maketitle
\begin{abstract}
   Localization Quality Estimation (LQE) is crucial and popular in the recent advancement of dense object detectors since it can provide accurate ranking scores that benefit the Non-Maximum Suppression processing and improve detection performance. As a common practice, most existing methods predict LQE scores through vanilla convolutional features shared with object classification or bounding box regression. In this paper, we explore a completely novel and different perspective to perform LQE -- based on the learned distributions of the four parameters of the bounding box. The bounding box distributions are inspired and introduced as ``General Distribution'' in GFLV1, which describes the uncertainty of the predicted bounding boxes well. Such a property makes the distribution statistics of a bounding box highly correlated to its real localization quality. Specifically, a bounding box distribution with a sharp peak usually corresponds to high localization quality, and vice versa. By leveraging the close correlation between distribution statistics and the real localization quality, we develop a considerably lightweight Distribution-Guided Quality Predictor (DGQP) for reliable LQE based on GFLV1, thus producing GFLV2. To our best knowledge, it is the first attempt in object detection to use a highly relevant, statistical representation to facilitate LQE. Extensive experiments demonstrate the effectiveness of our method. Notably, GFLV2 (ResNet-101) achieves 46.2 AP at 14.6 FPS, surpassing the previous state-of-the-art ATSS baseline (43.6 AP at 14.6 FPS) by absolute 2.6 AP on COCO {\tt test-dev}, without sacrificing the efficiency both in training and inference. Codes are available at https://github.com/implus/GFocalV2.

\end{abstract}

\begin{figure}[b]
    \vspace{-30pt}
	\begin{center}
		\setlength{\fboxrule}{0pt}
		\fbox{\includegraphics[width=0.48\textwidth]{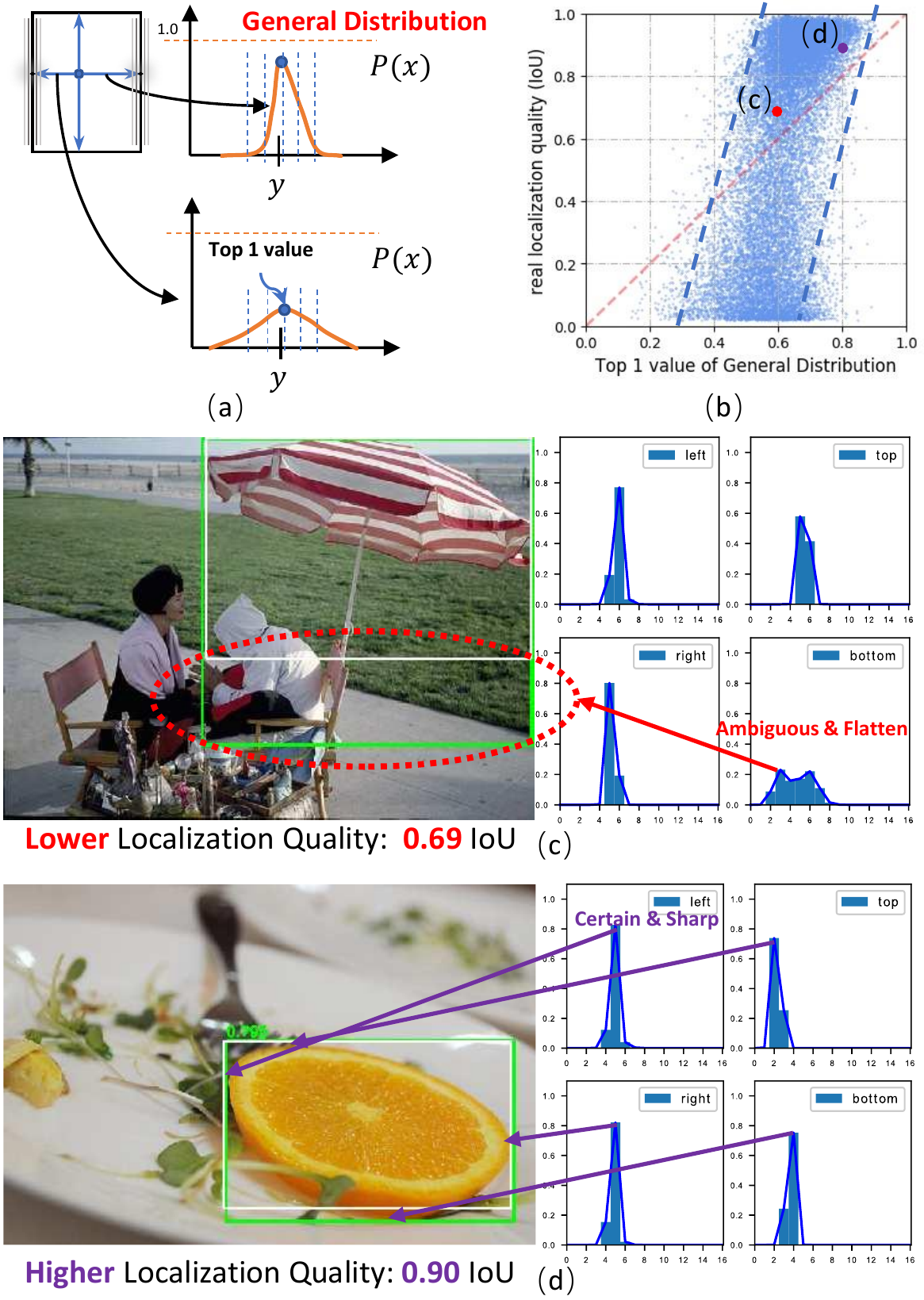}}
	\end{center}	
	\vspace{-20pt}
	\captionsetup{font={small}}
	\caption{Motivation of utilizing the highly relevant statistics of learned bounding box distributions to guide the better generation of its estimated localization quality. (a): The illustration of General Distribution in GFLV1 \cite{li2020generalized} to represent bounding boxes, which models the probability distribution of the predicted edges. (b): The scatter diagram of the relation between Top-1 (mean of four sides) value of General Distribution of predicted boxes and their real localization quality (IoU between the prediction and ground-truth), calculated over all validation images on COCO \cite{lin2014microsoft} dataset, based on GFLV1 model. (c) and (d): Two specific examples from (b), where the sharp distribution corresponds to higher quality, whilst the flat one stands for lower quality usually. Green: predicted bounding boxes; White: ground-truth bounding boxes.
	}
	\label{fig_concept}
	\vspace{-2pt}
\end{figure}

\begin{figure*}[t]
	\begin{center}
		\setlength{\fboxrule}{0pt}
		\fbox{\includegraphics[width=0.9\textwidth]{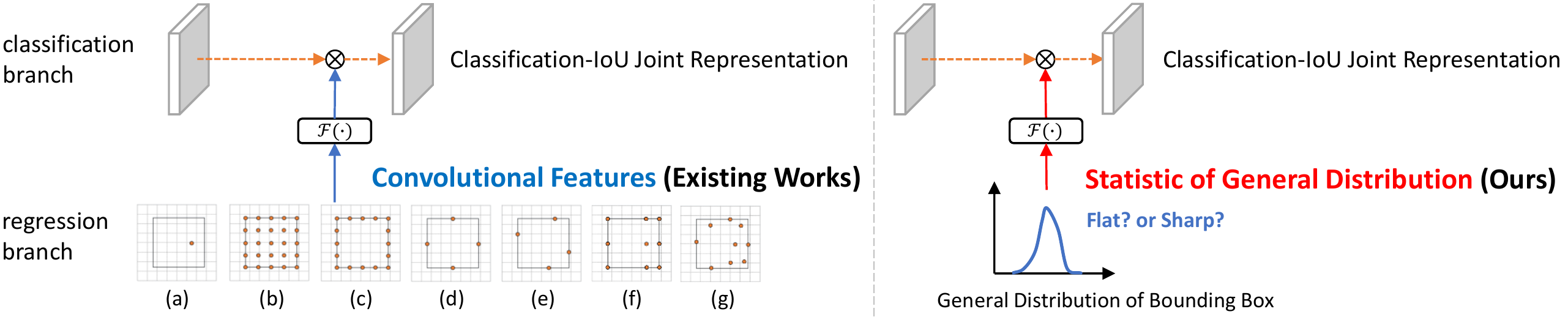}}
	\end{center}	
	\captionsetup{font={small}}
	\vspace{-18pt}
	\caption{Comparisons of input features for predicting localization quality between existing works (left) and ours (right). Existing works focus on different \emph{spatial locations} of convolutional features, including (a): \emph{point} \cite{redmon2016you,redmon2017yolo9000,redmon2018yolov3,tian2019fcos,zhang2019bridging,wu2020iou,kim2020probabilistic,li2020generalized}, (b): \emph{region} \cite{jiang2018acquisition}, (c): \emph{border} \cite{qiu2020borderdet} \emph{dense} points, (d): \emph{border} \cite{qiu2020borderdet} \emph{middle} points, (e): \emph{border} \cite{qiu2020borderdet} \emph{extreme} points, (f): \emph{regular} sampling points \cite{zhang2020varifocalnet}, and (g): \emph{deformable} sampling points \cite{chen2020reppoints,dai2017deformable}. In contrast, we use the statistic of learned box distribution to produce reliable localization quality.}
	\label{fig_comparison_concept}
	\vspace{-10pt}
\end{figure*}

\section{Introduction}
Dense object detector~\cite{redmon2016you,liu2016ssd,zhou2019objects,tian2019fcos,li2020generalized,qiu2020borderdet} which directly predicts pixel-level object categories and bounding boxes 
over feature maps, 
becomes increasingly popular due to its elegant and effective 
framework.
One of the crucial techniques underlying this framework is Localization Quality Estimation (LQE).
With the help of better LQE, high-quality bounding boxes tend to score higher than low-quality ones, greatly reducing the risk of mistaken suppression in Non-Maximum Suppression (NMS) processing.

Many previous works~\cite{redmon2016you,redmon2017yolo9000,redmon2018yolov3,tian2019fcos,zhang2019bridging,wu2020iou,kim2020probabilistic,li2020generalized,zhang2020varifocalnet,zhu2020autoassign,qiu2020borderdet} have explored LQE.
For example,
the YOLO family \cite{redmon2016you,redmon2017yolo9000,redmon2018yolov3} first adopt \emph{Objectness} to describe the localization quality, which is defined as the Intersection-over-Union (IoU) between the predicted and ground-truth box.
After that, \emph{IoU} is further explored and proved to be effective in IoU-Net \cite{jiang2018acquisition}, IoU-aware \cite{wu2020iou}, PAA \cite{kim2020probabilistic}, GFLV1 \cite{li2020generalized} and VFNet \cite{zhang2020varifocalnet}.
Recently, FCOS~\cite{tian2019fcos} and ATSS~\cite{zhang2019bridging} introduce
\emph{Centerness}, the distance degree to the object center, to suppress low-quality detection results. 
Generally, the aforementioned methods share a common characteristic that they are all based on \emph{vanilla convolutional
features}, e.g., features of points, borders or regions (see Fig.~\ref{fig_comparison_concept} (a)-(g)), to estimate the localization quality. 


Different from previous works,
in this paper, 
we explore 
a brand new 
perspective to conduct LQE -- by 
directly utilizing \emph{the statistics of bounding box distributions}, instead of using the \emph{vanilla convolutional features} (see Fig.~\ref{fig_comparison_concept}).
Here the \emph{bounding box distribution} is introduced as ``General Distribution'' in GFLV1~\cite{li2020generalized}, where it learns a discrete probability distribution of each predicted edge (Fig.~\ref{fig_concept} (a)) for describing the uncertainty of bounding box regression.
Interestingly, 
we 
observe that 
the statistic
of the General Distribution has a strong correlation with its real localization quality, as illustrated in Fig.~\ref{fig_concept} (b). More specifically in Fig.~\ref{fig_concept} (c) and (d), the shape (flatness) of bounding box distribution can clearly reflect the 
localization quality of the predicted results:
the sharper the distribution, the more accurate the predicted bounding box, and vice versa. Consequently, it can potentially be easier and very efficient to conduct better LQE by the guidance of the distribution information, as the input (distribution statistics of bounding boxes) and the output (LQE scores) are highly correlated.

Inspired by the strong correlation between the distribution statistics and LQE scores, we propose a very lightweight sub-network with only dozens of (e.g., 64) hidden units, on top of 
these distribution statistics to produce reliable LQE scores, significantly boosting the detection performance. Importantly, it brings negligible additional computation cost in practice and almost does not affect the training/inference speed of the basic object detectors. In this paper, we term this lightweight sub-network as Distribution-Guided Quality Predictor (DGQP), since it relies on the guidance of distribution statistics for quality predictions. 

By introducing the lightweight DGQP that predicts reliable LQE scores via statistics of bounding box distributions, we develop a novel dense object detector based on the framework of GFLV1, thus termed GFLV2.
To verify the effectiveness of GFLV2, we conduct extensive experiments on the challenging benchmark COCO~\cite{lin2014microsoft}.
Notably, based on ResNet-101 \cite{he2016deep}, GFLV2 achieves impressive detection performance (46.2 AP), i.e., 2.6 AP gains over the state-of-the-art ATSS baseline (43.6 AP) on COCO {\tt test-dev}, under the same training schedule and without sacrificing the efficiency both in training and inference.

In summary, our contributions are as follows:

$\bullet$ To our best knowledge, our work is the first to bridge the statistics of 
bounding box distributions and 
localization quality estimation in an end-to-end dense object detection 
framework.

$\bullet$ The proposed GFLV2 is considerably lightweight and cost-free in practice. It can also be easily plugged into most dense object detectors with a consistent gain of $\sim$2 AP, and without loss of training/inference speed.
	
$\bullet$ 
Our GFLV2 (Res2Net-101-DCN)
achieves very competitive 53.3 AP (multi-scale testing) on COCO dataset among dense object detectors.

\section{Related Work}
\noindent\textbf{Formats of LQE}: 
Early popular object detectors \cite{girshick2015fast,ren2015faster,cai2018cascade,he2017mask} 
simply treat the classification confidence as the formulation of LQE score,
but there is an obvious inconsistency between them, which inevitably degrades the detection performance.
To alleviate this problem, AutoAssign \cite{zhu2020autoassign} and BorderDet \cite{qiu2020borderdet} employ additional 
localization features to rescore the classification confidence, but they still lack an explicit definition of LQE. 

Recently, FCOS \cite{tian2019fcos} and ATSS \cite{zhang2019bridging} introduce a novel format of LQE, termed \emph{Centerness}, which depicts the distance degree to the center of the object.
Although \emph{Centerness} is 
effective,
recent researches \cite{li2020generalized,zhang2020varifocalnet} show that it has certain limitations and may be suboptimal for LQE.
SABL \cite{wang2019side} introduces boundary buckets for coarse localization, and utilizes the averaged bucketing confidence as a formulation of LQE.

After years of 
technical iterations~\cite{redmon2016you,redmon2017yolo9000,redmon2018yolov3,jiang2018acquisition,tychsen2018improving,huang2019mask,wu2020iou,kim2020probabilistic,li2020generalized,zhang2020varifocalnet}, \emph{IoU} has been deeply studied and becomes increasingly popular as an excellent measurement of LQE.
\emph{IoU} is first known as the \emph{Objectness} in YOLO \cite{redmon2016you,redmon2017yolo9000,redmon2018yolov3}, where the network is supervised to produce estimated IoUs between predicted boxes and ground-truth ones, to 
reduce ranking basis during NMS. 
Following the similar paradigm, IoU-Net \cite{jiang2018acquisition}, Fitness NMS \cite{tychsen2018improving}, MS R-CNN \cite{huang2019mask}, IoU-aware \cite{wu2020iou}, PAA \cite{kim2020probabilistic} utilize a separate branch to perform LQE in the \emph{IoU} form.
Concurrently, GFLV1 \cite{li2020generalized} and VFNet \cite{zhang2020varifocalnet} demonstrate a more effective format, by merging the classification score with \emph{IoU} to reformulate a joint representation.
Due to its great success \cite{li2020generalized,zhang2020varifocalnet}, we build our GFLV2 based on 
the Classification-IoU Joint Representation \cite{li2020generalized},
and develop a novel approach for reliable LQE.

\noindent\textbf{Input Features for LQE}: As shown in the left part of Fig.~\ref{fig_comparison_concept}, previous works directly use 
convolutional features as input for 
LQE, which only differ in the way of spatial sampling.
Most existing methods~\cite{redmon2016you,redmon2017yolo9000,redmon2018yolov3,tian2019fcos,zhang2019bridging,wu2020iou,kim2020probabilistic,li2020generalized} adopt the point features (see Fig.~\ref{fig_comparison_concept} (a)) to produce LQE scores for high efficiency. 
IoU-Net \cite{jiang2018acquisition} predicts IoU based on the region features as shown in Fig.~\ref{fig_comparison_concept} (b).
BorderDet \cite{qiu2020borderdet} designs three types of border-sensitive features (see Fig.~\ref{fig_comparison_concept} (c)-(e)) to facilitate LQE.
Similar with BorderDet, 
a star-shaped sampling manner (see Fig.~\ref{fig_comparison_concept} (f))
is designed in VFNet \cite{zhang2020varifocalnet}.
Alternatively, HSD \cite{cao2019hierarchical} and RepPoints \cite{yang2019reppoints,chen2020reppoints} focus on features with learned locations~(see Fig.~\ref{fig_comparison_concept} (g)) via the deformable convolution \cite{dai2017deformable,zhu2019deformable}.

The aforementioned methods mainly focus on extracting discriminating convolutional features with various spatial aspects for better LQE.
Different from previous methods, our proposed GFLV2 is designed in an
artful perspective: 
predicting LQE scores by its directly correlated variables---the statistics of bounding box distributions (see the right part of Fig.~\ref{fig_comparison_concept}).
As later demonstrated in 
Table~\ref{tab_input_source},
compared with convolutional features shown in Fig.~\ref{fig_comparison_concept} (a)-(g),
the statistics of bounding box distributions achieve an impressive efficiency and a high accuracy simultaneously.

\section{Method}
In this section, we first 
briefly review the
Generalized Focal Loss (i.e., GFLV1 \cite{li2020generalized}), and then derive the proposed GFLV2 based on the relevant 
concepts and formulations.

\subsection{Generalized Focal Loss V1}
\noindent \textbf{Classification-IoU Joint Representation}: 
This representation
is 
the key
component in GFLV1, which is designed to 
reduce the inconsistency between localization quality estimation and object classification during training and inference.
Concretely,
given an object with category label 
$c \in \{1, 2, ..., m\}$ 
($m$ 
indicates
the total number of categories),
GFLV1 utilizes the classification branch to produce the joint representation of Classification and IoU as $\textbf{J} = [J_1, J_2, ..., J_m]$, which 
satisfies:
\begin{equation}
\begin{split}
J_i = 
\left\{
\begin{array}{ll}
{\rm IoU}(b_{pred}, b_{gt}), & {\rm if\ }i=c; \\  
0, & {\rm otherwise},
\end{array}
\right.
\end{split}
\end{equation}
where ${\rm IoU}(b_{pred}, b_{gt})$ denotes the IoU between the predict bounding box $b_{pred}$ and the ground truth $b_{gt}$.

\noindent \textbf{General Distribution of Bounding Box Representation}: 
Modern detectors~\cite{ren2015faster,lin2017focal,tian2019fcos} usually describe the bounding box regression by Dirac delta distribution: $y = \int_{-\infty}^{+\infty} \delta(x - y) x \dif x$. 
Unlike them, GFLV1 introduces a flexible General Distribution $P(x)$ to represent the bounding box, where each edge of the bounding box can be formulated as: $\hat{y} = \int_{-\infty}^{+\infty} P(x) x \dif x = \int_{y_0}^{y_n} P(x) x \dif x$, under a predefined output range of $[y_0, y_n]$. 
To be compatible with the convolutional networks, the continuous domain is converted into the discrete one, via discretizing the range $[y_0, y_n]$ into a list $[y_0, y_1, ..., y_i, y_{i+1}, ..., y_{n-1}, y_n]$ with even intervals $\varDelta$ ($\varDelta = y_{i+1} - y_{i}, \forall i \in \{0, 1, ..., n-1\}$). As a result, given the discrete distribution property $\sum_{i = 0}^{n} P(y_i) = 1$, the estimated regression value $\hat{y}$ can be presented as:
\begin{equation}
\setlength{\abovedisplayskip}{2pt}
\setlength{\belowdisplayskip}{2pt}
\hat{y} = \sum_{i = 0}^{n} P(y_i) y_i.
\end{equation}

Compared with the Dirac delta distribution, the General Distribution $P(x)$ can faithfully reflect the prediction quality (see Fig.~\ref{fig_concept} (c)-(d)), which is the cornerstone of this work.

\begin{figure*}[t]
	\begin{center}
		\setlength{\fboxrule}{0pt}
		\fbox{\includegraphics[width=0.98\textwidth]{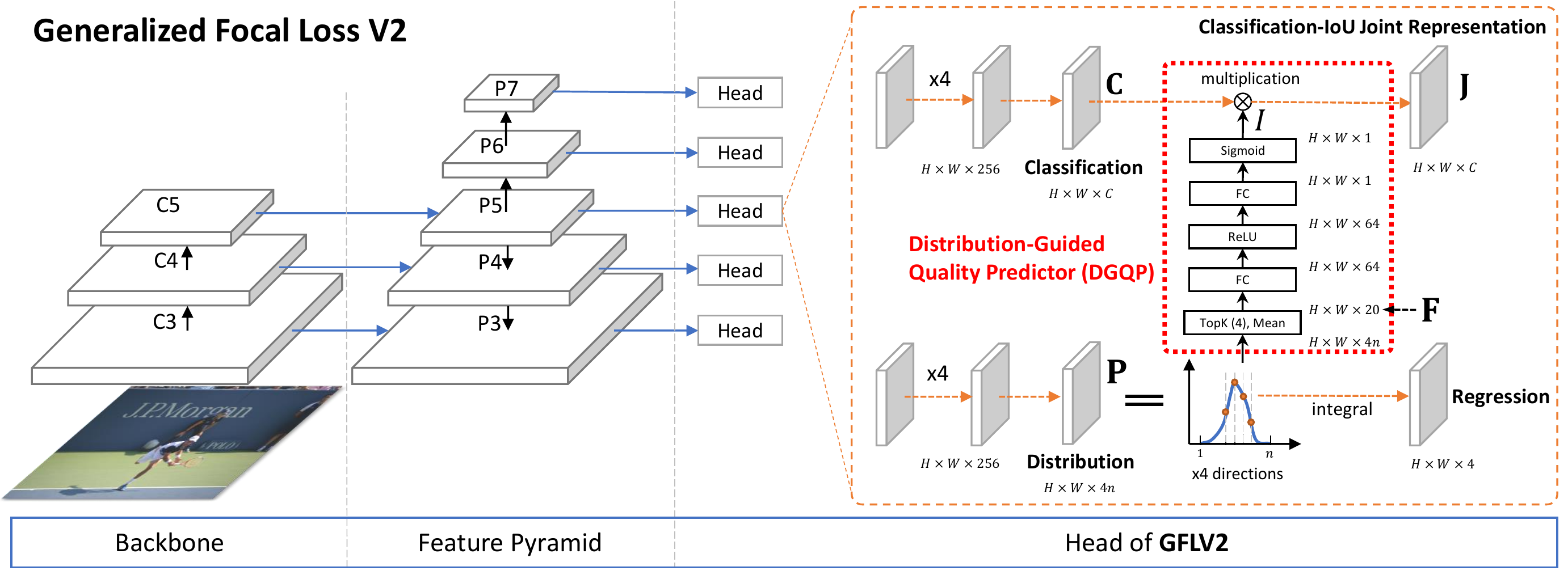}}
	\end{center}	
	\captionsetup{font={small}}
	\vspace{-20pt}
	\caption{The illustration of the proposed Generalized Focal Loss V2 (GFLV2), where a novel and tiny Distribution-Guided Quality Predictor (DGQP) uses the statistics of learned bounding box distributions to facilitate generating reliable IoU quality estimations.}
	\label{fig_method_cropped}
	\vspace{-10pt}
\end{figure*}
\subsection{Generalized Focal Loss V2}
\label{sec_gflv2}
\noindent 
\textbf{Decomposed Classification-IoU Representation}:
Although the joint representation solves 
the inconsistency problem~\cite{li2020generalized} between 
{object classification}
and 
{quality estimation during training and testing,}
there are still some limitations in using 
{only}
{the classification branch to predict the joint representation.}
In this work, we decompose the joint representation explicitly by leveraging information from both classification ($\textbf{C}$) and regression ($I$) branches:
\begin{equation}
\setlength{\abovedisplayskip}{2pt}
\setlength{\belowdisplayskip}{2pt}
{\textbf{J} = \textbf{C} \times {I},}
\label{jci}
\end{equation}
where $\textbf{C} = [C_1, C_2, ..., C_m], C_i \in [0, 1]$
denotes the Classification Representation of total $m$ categories, and ${I} \in [0, 1]$ is a scalar that stands for the IoU Representation.

Although $\textbf{J}$ is decomposed into two components, 
we use the final joint formulation (i.e., $\textbf{J}$) in both the training and testing phases,
so it can still avoid the inconsistency problem as mentioned in GFLV1.
Specifically, 
we first combine $\textbf{C}$ from the classification branch and ${I}$ from the proposed Distribution-Guided Quality Predictor (DGQP) in  regression branch, into the unified form $\textbf{J}$. Then, $\textbf{J}$ is supervised by Quality Focal loss (QFL) as proposed in~\cite{li2020generalized} during training, and used directly as NMS score in inference.

\noindent \textbf{Distribution-Guided Quality Predictor}: 
DGQP is the key component of GFLV2.
It delivers the statistics of the learned General Distribution $\mathbf{P}$ into a tiny sub-network (see \textcolor{red}{red} dotted frame in Fig.~\ref{fig_method_cropped}) to obtain the predicted IoU scalar $I$, which helps to generate high-quality Classification-IoU Joint Representation (Eq.~\eqref{jci}). 
Following GFLV1 \cite{li2020generalized},
we adopt the relative offsets from the location to the four sides of a bounding box as the regression targets, which are represented by the General Distribution. 
%
For convenience, we mark the left, right, top and bottom sides as $\{l, r, t, b\}$, 
and define the discrete 
probabilities
of the $w$ side as $\mathbf{P}^w = [P^w(y_0), P^w(y_1), ..., P^w(y_n)]$, where $w \in \{l, r, t, b\}$.

As illustrated in Fig.~\ref{fig_concept}, the flatness of the learned distribution is highly related to the quality of the final detected bounding box, and 
some relevant statistics can be used to reflect the flatness of the General Distribution.
As a result, such statistical features have a very strong correlation with the localization quality, which will ease the training difficulty and improves the quality of estimation. 
Practically, we recommand to choose the Top-$k$ values along with the mean value 
of each distribution vector $\mathbf{P}^w$, and concatenate them as the basic statistical feature $\mathbf{F} \in \mathbb{R}^{4(k+1)}$:
\begin{equation}
\setlength{\abovedisplayskip}{3pt}
\setlength{\belowdisplayskip}{3pt}
\mathbf{F} = {\rm Concat}\big(\big\{{\rm Topkm}(\mathbf{P}^w)\ |\ w \in \{l, r, t, b\}\big\}\big),
\end{equation}
where ${\rm Topkm}(\cdot)$ denotes the joint operation of calculating Top-$k$ values and their mean value. ${\rm Concat}(\cdot)$ means the channel concatenation. Selecting Top-$k$ values and their mean value as the input statistics have two benefits: 

$\bullet$ Since the sum of $\mathbf{P}^w$ is fixed (i.e., $\sum_{i=0}^{n}{P}^w(y_i) = 1$), Top-$k$ values along with their mean value can basically reflect the flatness of the distribution: the larger, the sharper; the smaller, the flatter;

\begin{figure}[h]
	\vspace{-10pt}
	\begin{center}
		\setlength{\fboxrule}{0pt}
		\fbox{\includegraphics[width=0.48\textwidth]{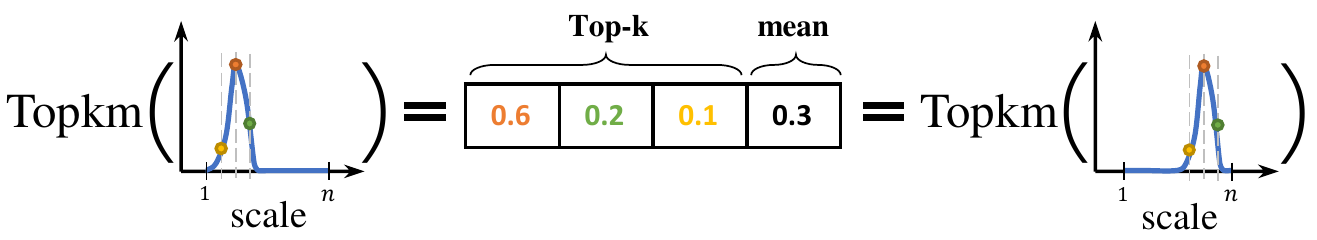}}
	\end{center}	
	\captionsetup{font={small}}
	\vspace{-20pt}
	\caption{$\text{{Topkm}}(\cdot)$ feature is robust to object scales.}
	\label{fig_topk_cropped}
	\vspace{-10pt}
\end{figure}

$\bullet$ Top-$k$ and mean values can make the statistical feature insensitive to its relative offsets over the distribution domain (see Fig.~\ref{fig_topk_cropped}), resulting in a robust representation which is not affected by object scales.

Given the statistical feature $\mathbf{F}$ of General Distribution as input, we design a very tiny sub-network $\mathcal{F}(\cdot)$ to predict the final IoU quality estimation. 
The sub-network has only two Fully-Connected (FC) layers, which are followed by ReLU \cite{krizhevsky2012imagenet} and Sigmoid, respectively. Consequently, the IoU scalar $I$ can be calculated as:
\begin{equation}
\setlength{\abovedisplayskip}{2pt}
\setlength{\belowdisplayskip}{2pt}
I = \mathcal{F}(\mathbf{F}) = \sigma(\mathbf{W}_2 \delta(\mathbf{W}_1 \mathbf{F})),
\end{equation}
where $\delta$ and $\sigma$ refer to the ReLU and Sigmoid, respectively. $\mathbf{W}_1 \in \mathbb{R}^{p\times 4(k+1)}$ and $\mathbf{W}_2 \in \mathbb{R}^{1\times p}$. $k$ denotes the Top-$k$ parameter and $p$ is the channel dimension of the hidden layer ($k$ = 4, $p$ = 64 is a typical setting in our experiment).

\noindent \textbf{Complexity}: The overall architecture of GFLV2 is illustrated in Fig.~\ref{fig_method_cropped}. It is worth noting that the DGQP module is very lightweight. First, it only brings
thousands of
additional parameters, which are negligible compared to the number of parameters of the entire detection model.
For example, for the model with ResNet-50 \cite{he2016deep} and FPN \cite{lin2017feature}, the extra parameters of the DGQP module only account for $\sim$0.003\%. 
Second, 
the computational overhead of the DGQP module is also very small due to its extremely light structure.
As shown in Table~\ref{tab_compatibility} and \ref{tab_efficiency}, the use of the DGQP module hardly reduces the training and inference speed of the original detector in practice.

\begin{table}[t]
	\small
	\centering
	\renewcommand\arraystretch{1.1}
	\newcommand{\tabincell}[2]{\begin{tabular}{@{}#1@{}}#2\end{tabular}}
	
	\resizebox{0.46\textwidth}{!}{
		\begin{tabular}{c|c|c|c| c c c}
			\hline
			Top-$k$ (4) & Mean & Var & Dim & AP & AP$_{50}$ & AP$_{75}$  \\
			\hline\hline
			$\checkmark$ &              &              & 16 & 40.8 & 58.5 & 44.2\\
			& $\checkmark$ &              & 4  & 40.2 & 58.5 & 43.6\\
			&              & $\checkmark$ & 4  & 40.3 & 58.3 & 43.7\\
			$\checkmark$ & $\checkmark$ &              & 20 & \textbf{41.1} & \textbf{58.8} & \textbf{44.9} \\
			$\checkmark$ &              & $\checkmark$ & 20 & 40.9 & 58.5 & 44.7\\
			$\checkmark$ & $\checkmark$ & $\checkmark$ & 24 & 40.9 & 58.4 & 44.7 \\
			\hline
		\end{tabular}
	}
	\captionsetup{font={small}}
	\vspace{-8pt}
	\caption{Performances of different combinations of the input statistics by fixing $k = 4$ and $p = 64$. ``Mean'' denotes the mean value, ``Var'' denotes the variance number, and ``Dim'' is short for ``Dimension'' that means the total amount of the input channels.}
	\vspace{-10pt}
	\label{tab_input_statistics}
\end{table}

\section{Experiment}

\noindent \textbf{Experimental Settings}: We conduct experiments on COCO benchmark \cite{lin2014microsoft}, where {\tt  trainval35k} with 115K images is used for training and {\tt minival} with 5K images for validation in our ablation study. Besides, we obtain the main results on {\tt test-dev} with 20K images from the evaluation server. All results are produced under mmdetection \cite{chen2019mmdetection} for fair comparisons, where the default hyper-parameters are always adopted. Unless otherwise stated, we apply the standard 1x learning schedule (12 epochs) without multi-scale training for the ablation study, based on ResNet-50 \cite{he2016deep} backbone. The training/testing details follow the descriptions in previous works \cite{li2020generalized,chen2020reppoints}.



\subsection{Ablation Study}
\noindent \textbf{Combination of Input Statistics}: In addition to the pure Top-$k$ values, there are some statistics that may reflect more characteristics of the distributions, such as the mean and variance of these Top-$k$ numbers. Therefore, we conduct experiments to investigate the effect of their combinations as input, by fixing $k = 4$ and $p = 64$. From Table~\ref{tab_input_statistics}, we observe that the Top-$4$ values with their mean number perform best. Therefore, we default to use such a combination as the standard statistical input in the following experiments.

\noindent \textbf{Structure of DGQP (i.e., $k$, $p$)}: We then examine the impact of different parameters of $k$, $p$ in DGQP on the detection performance. Specifically, we report the effect of $k$ and $p$ by fixing one and varying another in Table~\ref{tab_kp}. It is observed that $k = 4, p = 64$ steadily achieves the optimal accuracy among various combinations.

\begin{table}[t]
	\small
	\centering
	\renewcommand\arraystretch{1.1}
	\newcommand{\tabincell}[2]{\begin{tabular}{@{}#1@{}}#2\end{tabular}}
	
	\resizebox{0.46\textwidth}{!}{
		\begin{tabular}{c|c|c c c|c c c}
			\hline
			$k$ & $p$ & AP & AP$_{50}$ & AP$_{75}$ & AP$_{S}$ & AP$_{M}$ & AP$_{L}$  \\
			\hline\hline
			0  &-- & {40.2} & {58.6} & 43.4 & 23.0 & {44.3} & {53.0} \\\hline
			1  & \multirow{6}{*}{64} & 40.2 & 58.3 & 44.0 & 23.4 & 44.1 & 52.1 \\
			2  & & 40.9 & 58.5 & 44.6 & 23.3 & 44.8 & \textbf{53.5}\\
			3  & & 40.9 & 58.5 & 44.6 & \textbf{24.3} & \textbf{44.9} & 52.3\\
			4  & & \textbf{41.1} & \textbf{58.8} & \textbf{44.9} & {23.5} & \textbf{44.9} & 53.3 \\
			8  & & 41.0 & 58.6 & 44.5 & 23.5 & 44.5 & 53.4 \\
			16 & & 40.8 & 58.5 & 44.4 & 23.4 & 44.2 & 53.1\\
			\hline
			\multirow{6}{*}{4} & 8 & 40.9 & 58.4 & 44.5 & 23.1 & 44.5 & 52.6\\
			  & 16  & 40.8 & 58.3 & 44.1 & 23.3 & 44.6 & 52.0 \\
			  & 32  & 40.9 & 58.7 & 44.3 & 23.1 & 44.6 & 53.2\\
			  & 64  & \textbf{41.1} & \textbf{58.8} & \textbf{44.9} & \textbf{23.5} & \textbf{44.9} & \textbf{53.3} \\
			  & 128 & 40.9 & 58.3 & 44.6 & 23.2 & 44.4 & 52.7 \\
			  & 256 & 40.7 & 58.3 & 44.4 & 23.4 & 44.3 & 52.9\\
			\hline
		\end{tabular}
	}
\captionsetup{font={small}}
	\vspace{-8pt}
	\caption{Performances of various $k, p$ in DGQP. $k = 0$ denotes the baseline version without the usage of DGQP (i.e., GFLV1).}
	\vspace{-6pt}
	\label{tab_kp}
\end{table}

\noindent \textbf{Type of Input Features}: To the best of our knowledge, the proposed DGQP is the first to use the statistics of learned distributions of bounding boxes for the generation of better LQE scores in the literature. Since the input (distribution statistics) and the output (LQE scores) are highly correlated, we speculate that it can be more effective or efficient than ordinary convolutional input proposed in existing methods. Therefore, we fix the hidden layer dimension of DGQP (i.e., $p=64$) and compare our statistical input with most existing possible types of convolutional inputs, from point (a), region (b), border (c)-(e), regular points (f), and deformable points (g), respectively (Fig.~\ref{fig_comparison_concept}). Table~\ref{tab_input_source} shows that our distribution statistics perform best in overall AP, also fastest in inference, compared against various convolutional features.

\begin{table}[t]
	\small
	\centering
	\renewcommand\arraystretch{1.1}
	\newcommand{\tabincell}[2]{\begin{tabular}{@{}#1@{}}#2\end{tabular}}
	
	\resizebox{0.48\textwidth}{!}{
		\begin{tabular}{c|c| c c c | c}
			\hline
			\multicolumn{2}{c|}{Input Feature} & AP & AP$_{50}$ & AP$_{75}$ & FPS \\
			\hline\hline
			\multicolumn{2}{c|}{ Baseline (ATSS \cite{zhang2019bridging} w/ QFL \cite{li2020generalized})} & {39.9} & {58.5} & {43.0} & \textbf{19.4}\\
			\hline
			\multirow{7}{*}{\tabincell{c}{Convolutional Features}} & (a) &  40.2 & 58.6 & 43.7 & {19.3} \\
			 & (b) & 40.5 & 59.0 & 44.0 & 14.0\\ 
			 & (c) & 40.5 & 58.7 & 44.1 & 16.2\\
			 & (d) & 40.6 & 59.0 & 44.0 & 18.3\\
			 & (e) & 40.6 & 58.9 & 44.1 & 17.8\\
			 & (f) & 40.7 & 59.0 & 44.1 & 17.9 \\
			 & (g) & 40.8 & 58.9 & 44.6 & 18.4 \\\hline
			\multicolumn{2}{c|}{Distribution Statistics \textbf{(ours)} }  &  \textbf{41.1} & 58.8 & 44.9 & \textbf{19.4}\\
			\hline
		\end{tabular}
	}
\captionsetup{font={small}}
	\vspace{-8pt}
	\caption{Comparisons among different input features by fixing the hidden layer dimension of DGQP.}
	\vspace{-10pt}
	\label{tab_input_source}
\end{table}

\begin{figure}[t]
	\begin{center}
		\setlength{\fboxrule}{0pt}
		\fbox{\includegraphics[width=0.40\textwidth]{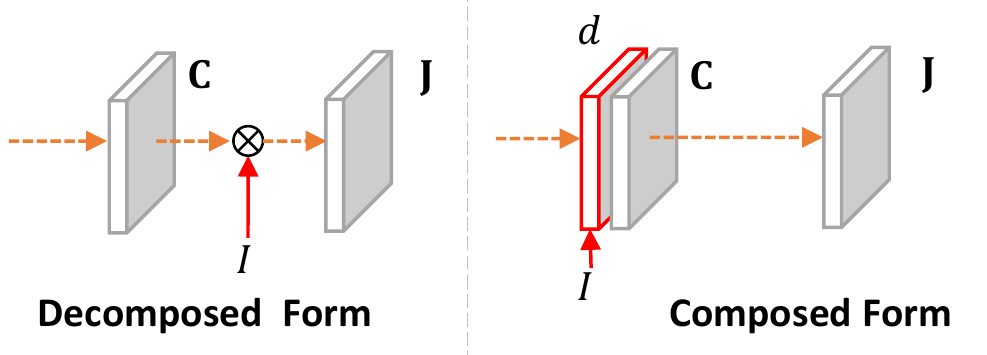}}
	\end{center}	
	\captionsetup{font={small}}
	\vspace{-18pt}
	\caption{Different ways to utilize the distribution statistics, including Decomposed Form (left) and Composed Form (right).}
	\label{fig_decompose_compare_cropped}
	\vspace{-10pt}
\end{figure}

\noindent \textbf{Usage of the Decomposed Form}: Next, we examine what is the best formulation of Classification-IoU Joint Representation in the case of using distribution statistics. There are basically two formats: the Composed Form and the proposed Decomposed Form (Sec.~\ref{sec_gflv2}), as illustrated in Fig.~\ref{fig_decompose_compare_cropped}. Here the ``Decomposed'' (the left part of Fig.~\ref{fig_decompose_compare_cropped}) means that the final joint representation can be explicitly decomposed through multiplication by two components, i.e., $\textbf{J} = \textbf{C} \times {I}$ in Eq.~\eqref{jci}. Whilst the ``Composed'' (the right part of Fig.~\ref{fig_decompose_compare_cropped}) shows that $\textbf{J}$ is directly obtained through FC layers where its input feature is enriched ($d$ is the dimension of the appended feature) by the information of distribution statistics. From Table~\ref{tab_decompose_compare}, our proposed Decomposed Form is always superior than the Composed Forms with various $d$ settings in both accuracy and running speed.

\begin{table}[t]
	\small
	\centering
	\renewcommand\arraystretch{1.1}
	\newcommand{\tabincell}[2]{\begin{tabular}{@{}#1@{}}#2\end{tabular}}
	
	\resizebox{0.46\textwidth}{!}{
		\begin{tabular}{c|l|c c c|c}
			\hline
			\multicolumn{2}{c|}{ Type} & AP & AP$_{50}$ & AP$_{75}$ & FPS \\
			\hline\hline
			\multicolumn{2}{c|}{ Baseline (GFLV1 \cite{li2020generalized}) } & {40.2} & {58.6} & 43.4 & \textbf{19.4} \\
			\hline
			\multirow{5}{*}{Composed Form} & $d=16$  & 40.5 & 58.5 & 43.7 & 19.2 \\ 
										   & $d=32$  & 40.5 & 58.5 & 43.7 & 19.2 \\
										   & $d=64$  & 40.7 & 58.5 & 44.3 & 19.1 \\
										   & $d=128$ & 40.7 & 58.6 & 44.4 & 18.9 \\
										   & $d=256$ & 40.7 & 58.3 & 44.2 & 18.5 \\\hline
			\multicolumn{2}{c|}{Decomposed Form \textbf{(ours)}} & \textbf{41.1} & \textbf{58.8} & \textbf{44.9} & \textbf{19.4} \\
			\hline
		\end{tabular}
	}
\captionsetup{font={small}}
	\vspace{-8pt}
	\caption{Comparisons between Decomposed Form (proposed) and Composed Form (with various dimension $d$ settings).}
	\vspace{-6pt}
	\label{tab_decompose_compare}
\end{table}

\noindent \textbf{Compatibility for Dense Detectors}: Since GFLV2 is very lightweight and can be adapted to various types of dense detectors, we employ it to a series of recent popular detection methods. For those detectors that do not support the distributed representation of bounding boxes, we make the minimal and necessary modifications to enable it to generate distributions for each edge of a bounding box. 
Based on the results in Table~\ref{tab_compatibility}, GFLV2 can consistently improve $\sim$2 AP in popular dense detectors, without loss of inference speed.

\begin{table}[t]
	\small
	\centering
	\renewcommand\arraystretch{1.1}
	\newcommand{\tabincell}[2]{\begin{tabular}{@{}#1@{}}#2\end{tabular}}
	
	\resizebox{0.48\textwidth}{!}{
		\begin{tabular}{c|c|l c c|c}
			\hline
			Method & GFLV2 & AP & AP$_{50}$ & AP$_{75}$ & FPS \\
			\hline\hline
			RetinaNet \cite{lin2017focal} & & 36.5 & 55.5 & 38.7 & 19.0\\
			RetinaNet \cite{lin2017focal}& $\checkmark$ & \textbf{38.6 {\scriptsize(+2.1)}} & \textbf{56.2} & \textbf{41.7} & 19.0\\ 
			\hline
			FoveaNet \cite{kong2019foveabox}& & 36.4 & 55.8 & 38.8 & 20.0\\
			FoveaNet \cite{kong2019foveabox}& $\checkmark$ & \textbf{38.5 {\scriptsize(+2.1)}} & \textbf{56.8} & \textbf{41.6} & 20.0\\ 
			\hline
			FCOS \cite{tian2019fcos} & & 38.5 & 56.9 & 41.4 & 19.4 \\
			FCOS \cite{tian2019fcos} & $\checkmark$ & \textbf{40.6 {\scriptsize(+2.1)}} & \textbf{58.2} & \textbf{43.9} & 19.4\\ 
			\hline
			ATSS \cite{zhang2019bridging}& & 39.2 & 57.4 & 42.2 & 19.4 \\
			ATSS \cite{zhang2019bridging}& $\checkmark$ & \textbf{41.1 {\scriptsize(+1.9)}} & \textbf{58.8} & \textbf{44.9} & 19.4\\ 
			\hline
		\end{tabular}
	}
\captionsetup{font={small}}
	\vspace{-8pt}
	\caption{Integrating GFLV2 into various popular dense object detectors. A consistent $\sim$2 AP gain is observed without loss of inference speed.} 
	\vspace{-6pt}
	\label{tab_compatibility}
\end{table}


\begin{figure}[t]
    \vspace{-8pt}
	\begin{center}
		\setlength{\fboxrule}{0pt}
		\fbox{\includegraphics[width=0.42\textwidth]{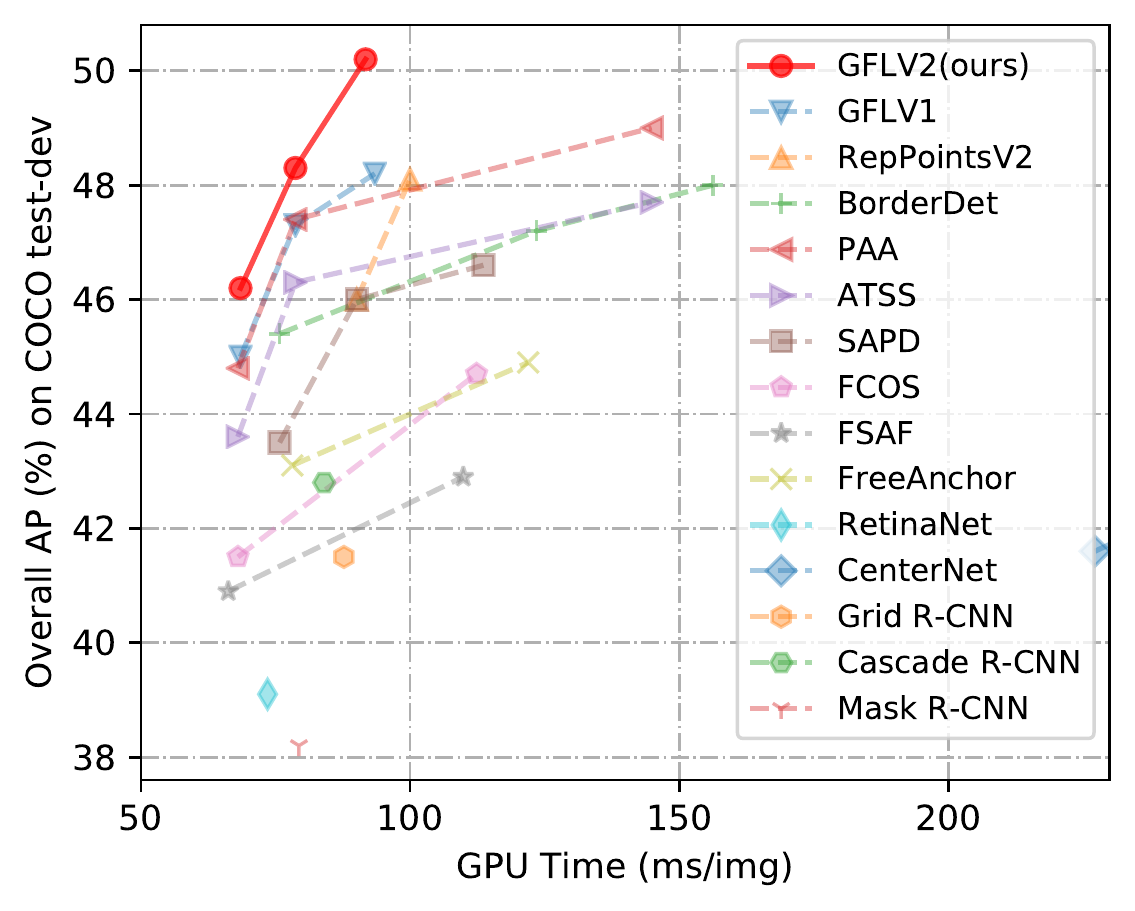}}
	\end{center}	
	\vspace{-24pt}
	\captionsetup{font={small}}
	\caption{Single-model single-scale speed (ms) vs. accuracy (AP) on COCO {\tt test-dev} among state-of-the-art approaches. GFLV2 achieves better speed-accuracy trade-off than its competitive counterparts.}
	\label{fig_performance}
	\vspace{-6pt}
\end{figure}

\subsection{Comparisons with State-of-the-arts}
In this section, we compare GFLV2 with state-of-the-art approaches on COCO {\tt test-dev} in Table~\ref{tab_sota}. Following previous works \cite{lin2017focal,tian2019fcos}, the multi-scale  ([480, 960]) training strategy and 2x learning schedule (24 epochs) are adopted during training. For a fair comparison, the results of single-model single-scale testing for all methods are reported, including their corresponding inference speeds (FPS). We also report additional multi-scale testing results for GFLV2. The visualizaion of the accuracy-speed trade-off is demonstrated in Fig.~\ref{fig_performance}, and we observe that GFLV2 pushes the envelope of accuracy-speed boundary to a new level. Our best result with a single Res2Net-101-DCN model achieves considerably competitive 53.3 AP. 
\begin{table}[t]
	\small
	\centering
	\renewcommand\arraystretch{1.1}
	\newcommand{\tabincell}[2]{\begin{tabular}{@{}#1@{}}#2\end{tabular}}
	\resizebox{0.4\textwidth}{!}{
		\begin{tabular}{l|l|c|l}
			\hline
			Method & AP & FPS & \tabincell{c}{PCC} $\uparrow$  \\
			\hline\hline
			FCOS$^*$ \cite{tian2019fcos}  & 39.1 & 19.4 & 0.624 \\
			ATSS$^*$ \cite{zhang2019bridging} & 39.9 & 19.4 & 0.631 \\\hline
			GFLV1    \cite{li2020generalized} & 40.2 & 19.4 & 0.634 \\ 
			
			GFLV2 \textbf{(ours)} & \textbf{41.1 {\scriptsize(+0.9)}} & 19.4 & \textbf{0.660 {\scriptsize(+0.26)}} \\ 
			\hline
		\end{tabular}
	}
	\captionsetup{font={small}}
	\vspace{-8pt}
	\caption{Pearson Correlation Coefficients (PCC) for representative dense object detectors. $^*$ denotes the application of Classification-IoU Joint Representation, instead of additional \emph{Centerness} branch. } 
	\vspace{-10pt}
	\label{tab_error}
\end{table}

\begin{table*}[t]
	\renewcommand\arraystretch{1.2}
	\centering
	\footnotesize \setlength{\tabcolsep}{4.5pt}
	\vspace{-2pt}
	\begin{threeparttable}
		\resizebox{\textwidth}{!}{
			\begin{tabular}{l|l|c|c|c|ccc|ccc|c}
				\hline
				Method  & Backbone & Epoch & MS$_{\text{train}}$ & FPS & AP &AP$_{50}$ &AP$_{75}$ &AP$_{\it S}$ &AP$_{\it M}$ &AP$_{\it L}$ & Reference\\
				\hline
				\textit{multi-stage:}  & & & & & & & & & & & \\
				Faster R-CNN w/ FPN \cite{lin2017feature}  & R-101 & 24 & & 14.2 & 36.2 & 59.1 & 39.0 & 18.2 & 39.0 & 48.2 & CVPR17 \\
				Cascade R-CNN \cite{cai2018cascade}  & R-101 & 18 & & 11.9 &42.8 &62.1 &46.3 &23.7 &45.5 &55.2 & CVPR18\\
				Grid R-CNN \cite{lu2019grid} & R-101 & 20 &  & 11.4 &41.5 & 60.9 & 44.5 & 23.3 & 44.9 & 53.1 & CVPR19 \\
				Libra R-CNN \cite{pang2019libra} & X-101-64x4d & 12 & & 8.5 & 43.0 & 64.0 & 47.0 & 25.3 & 45.6 & 54.6 & CVPR19\\
				RepPoints \cite{yang2019reppoints}   &R-101 & 24 & & 13.3 & 41.0 & 62.9 & 44.3 & 23.6 & 44.1 & 51.7 & ICCV19\\
				RepPoints \cite{yang2019reppoints}   &R-101-DCN& 24 & $\checkmark$ & 11.8 & 
				45.0 & 66.1 & 49.0 & 26.6 & 48.6 & 57.5 & ICCV19 \\
				RepPointsV2 \cite{chen2020reppoints} & R-101 & 24 & $\checkmark$ & 11.1 & 46.0 & 65.3 & 49.5 & 27.4 & 48.9 & 57.3 & NeurIPS20 \\
				RepPointsV2 \cite{chen2020reppoints} & R-101-DCN & 24 & $\checkmark$ & 10.0 &  48.1 & 67.5 & 51.8 & 28.7 & 50.9 & 60.8 & NeurIPS20 \\
				TridentNet \cite{li2019scale}& R-101 & 24 & $\checkmark$ & 2.7$^*$ & 42.7 & 63.6 & 46.5 & 23.9 & 46.6 & 56.6 & ICCV19 \\
				TridentNet \cite{li2019scale}& R-101-DCN & 36 & $\checkmark$ & 1.3$^*$ & 46.8 & 67.6 & 51.5 & 28.0 & 51.2 & 60.5 & ICCV19 \\
				TSD \cite{song2020revisiting} & R-101 & 20 &  & 1.1 & 43.2 & 64.0 & 46.9 & 24.0 & 46.3 & 55.8 & CVPR20\\
				BorderDet \cite{qiu2020borderdet} & R-101 & 24 & $\checkmark$ & 13.2$^*$ &  45.4 & 64.1 & 48.8 & 26.7 & 48.3 & 56.5 & ECCV20\\
				BorderDet \cite{qiu2020borderdet} & X-101-64x4d & 24 & $\checkmark$ & 8.1$^*$ &  47.2 & 66.1 & 51.0 & 28.1 & 50.2 & 59.9 & ECCV20\\
				BorderDet \cite{qiu2020borderdet} & X-101-64x4d-DCN & 24 & $\checkmark$ & 6.4$^*$ &  48.0 & 67.1 & 52.1 & 29.4 & 50.7 & 60.5 & ECCV20\\
				\hline
				\hline
				\textit{one-stage:}   & & & & & & & & & & & \\
				CornerNet \cite{law2018cornernet} &HG-104 & 200 & $\checkmark$ & 3.1$^*$ & 40.6 & 56.4 & 43.2 & 19.1 & 42.8 & 54.3 & ECCV18 \\
				CenterNet \cite{duan2019centernet} &HG-104 & 190 &  $\checkmark$ & 3.3$^*$ &44.9 &62.4 &48.1 &25.6 &47.4 &57.4 & ICCV19 \\
				CentripetalNet \cite{dong2020centripetalnet} & HG-104 & 210 & $\checkmark$ & n/a  & 45.8 & 63.0 & 49.3 & 25.0 & 48.2 & 58.7 & CVPR20\\
				RetinaNet \cite{lin2017focal} &R-101 &  18 &  & 13.6 &39.1 &59.1 &42.3 &21.8 &42.7 &50.2 & ICCV17 \\
				FreeAnchor \cite{zhang2019freeanchor} & R-101 & 24 & $\checkmark$ & 12.8 & 43.1 & 62.2 & 46.4 & 24.5 & 46.1 & 54.8 & NeurIPS19 \\
				FreeAnchor \cite{zhang2019freeanchor} & X-101-32x8d & 24 & $\checkmark$ & 8.2 & 44.9 & 64.3 & 48.5 & 26.8 & 48.3 & 55.9 & NeurIPS19 \\
				FSAF \cite{zhu2019feature} & R-101 & 18 & $\checkmark$ & 15.1 & 40.9 & 61.5 & 44.0 & 24.0 & 44.2 & 51.3 & CVPR19\\
				FSAF \cite{zhu2019feature}& X-101-64x4d & 18 & $\checkmark$ & 9.1 & 42.9 & 63.8 & 46.3 & 26.6 & 46.2 & 52.7 & CVPR19 \\
				FCOS \cite{tian2019fcos} & R-101 & 24 & $\checkmark$ & 14.7 &41.5 & 60.7 & 45.0 & 24.4 & 44.8 & 51.6 & ICCV19\\
				FCOS \cite{tian2019fcos} & X-101-64x4d & 24 & $\checkmark$ & 8.9 & 44.7 & 64.1 & 48.4 & 27.6 & 47.5 & 55.6 & ICCV19\\
				SAPD \cite{zhu2019soft}& R-101 & 24 & $\checkmark$ & 13.2 & 43.5 & 63.6 & 46.5 & 24.9 & 46.8 & 54.6 & CVPR20 \\
				SAPD \cite{zhu2019soft}&  R-101-DCN & 24 & $\checkmark$ & 11.1 & 46.0 & 65.9 & 49.6 & 26.3 & 49.2 & 59.6 & CVPR20 \\
				SAPD \cite{zhu2019soft}& X-101-32x4d-DCN & 24 & $\checkmark$ & 8.8 & 46.6 & 66.6 & 50.0 & 27.3 & 49.7 & 60.7 & CVPR20 \\
				ATSS \cite{zhang2019bridging}& R-101 & 24 & $\checkmark$ & 14.6 &43.6 &62.1 &47.4 &26.1 &47.0 &53.6 & CVPR20 \\
				ATSS \cite{zhang2019bridging} &R-101-DCN & 24 &  $\checkmark$ & 12.7 & 46.3 &64.7 &50.4 &27.7 &49.8 &58.4 & CVPR20\\
				ATSS \cite{zhang2019bridging}& X-101-32x8d-DCN & 24 & $\checkmark$ & 6.9 & {47.7} & {66.6} &{52.1} & {29.3} &50.8 & {59.7} & CVPR20 \\
				PAA \cite{kim2020probabilistic} & R-101 & 24 & $\checkmark$ & 14.6 & 44.8 & 63.3 & 48.7 & 26.5 & 48.8 & 56.3 & ECCV20\\
				PAA \cite{kim2020probabilistic} & R-101-DCN & 24 & $\checkmark$ & 12.7 & 47.4 & 65.7 & 51.6 & 27.9 & 51.3 & 60.6 & ECCV20\\
				PAA \cite{kim2020probabilistic} & X-101-64x4d-DCN & 24 & $\checkmark$ & 6.9 & 49.0 & 67.8 & 53.3 & 30.2 & 52.8 & 62.2 & ECCV20\\
				GFLV1 \cite{li2020generalized} &R-50 & 24 & $\checkmark$ & 19.4 & 43.1 & 62.0 & 46.8 & 26.0 & 46.7 & 52.3 & NeurIPS20\\
				GFLV1 \cite{li2020generalized}&R-101 & 24 & $\checkmark$ & 14.6 & 45.0 & 63.7 & 48.9 & 27.2 & 48.8 & 54.5 & NeurIPS20 \\
				GFLV1 \cite{li2020generalized}&R-101-DCN & 24 & $\checkmark$ & 12.7 & 47.3 & 66.3 & 51.4 & 28.0 & {51.1} & 59.2& NeurIPS20 \\
				GFLV1 \cite{li2020generalized}&X-101-32x4d-DCN & 24 &  $\checkmark$ & 10.7 & {48.2} & {67.4} & {52.6} & {29.2} & {51.7} & {60.2}& NeurIPS20 \\
				\hline
				GFLV2 \textbf{(ours)}&R-50 & 24 & $\checkmark$ & 19.4 & 44.3 & 62.3 & 48.5 &  26.8 & 47.7 & 54.1 & -- \\
				GFLV2 \textbf{(ours)}&R-101 & 24 & $\checkmark$ & 14.6 & 46.2 & 64.3 & 50.5 & 27.8 & 49.9 & 57.0 & -- \\
				GFLV2 \textbf{(ours)}&R-101-DCN & 24 & $\checkmark$ & 12.7 & 48.3 & 66.5 & 52.8 & 28.8 & 51.9 & 60.7 & -- \\
				GFLV2 \textbf{(ours)}&X-101-32x4d-DCN & 24 &  $\checkmark$ & 10.7 & 49.0 & 67.6 & 53.5 & 29.7 & 52.4 & 61.4 & -- \\
				GFLV2 \textbf{(ours)}&R2-101-DCN & 24 &  $\checkmark$ & 10.9 & 50.6 & 69.0 & 55.3 & 31.3 & 54.3 & 63.5 & -- \\
				GFLV2 \textbf{(ours)} + MS$_{\text{test}}$ &R2-101-DCN & 24 &  $\checkmark$ & -- & 53.3 & 70.9 & 59.2 & 35.7 & 56.1 & 65.6 & -- \\
				\hline
			\end{tabular}
		}
	\end{threeparttable}
    \captionsetup{font={small}}
	\vspace{-8pt}
	\caption{Comparisons between state-of-the-art detectors \emph{(single-model and single-scale results except the last row)} on COCO {\tt test-dev}. ``MS$_{\text{train}}$'' and ``MS$_{\text{test}}$'' denote multi-scale training and testing, respectively. FPS values with $^*$ are from \cite{zhu2019soft} or their official repositories \cite{qiu2020borderdet}, while others are measured on the same machine with a single GeForce RTX 2080Ti GPU under the same mmdetection \cite{chen2019mmdetection} framework, using a batch size of 1 whenever possible. ``n/a'' means that both trained models and timing results from original papers are not available. \textbf{R}: ResNet \cite{he2016deep}. \textbf{X}: ResNeXt \cite{xie2017aggregated}. \textbf{HG}: Hourglass \cite{newell2016stacked}. \textbf{DCN}: Deformable Convolutional Network \cite{zhu2019deformable}. \textbf{R2}: Res2Net \cite{gao2019res2net}.}
	\vspace{-6pt}
	\label{tab_sota}
\end{table*}

\begin{figure*}[t]
	\begin{center}
		\setlength{\fboxrule}{0pt}
		\fbox{\includegraphics[width=\textwidth]{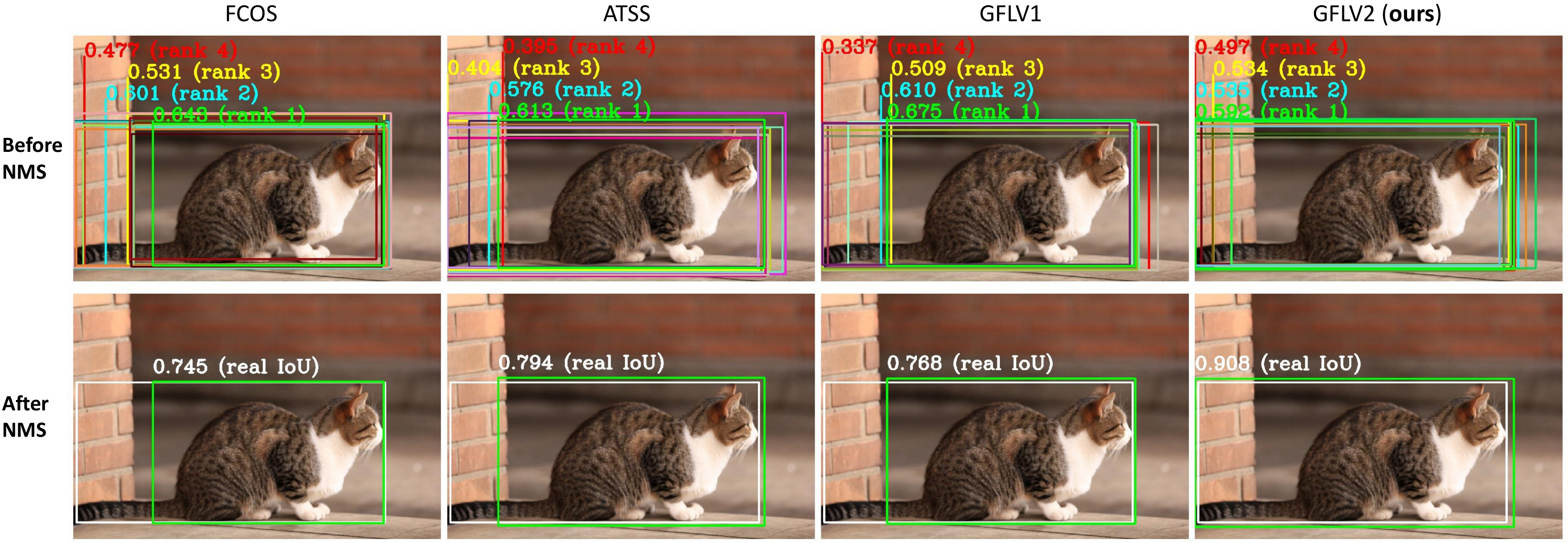}}
	\end{center}
	\captionsetup{font={small}}
	\vspace{-22pt}
	\caption{Visualization of predicted bounding boxes before and after NMS, along with their corresponding predicted LQE scores (only Top-4 scores are plotted for a better view). For many existing approaches \cite{tian2019fcos,zhang2019bridging,li2020generalized}, they fail to produce the highest LQE scores for the best candidates. In contrast, our GFLV2 reliably assigns larger quality scores for those real high-quality ones, thus reducing the risk of mistaken suppression in NMS processing. White: ground-truth bounding boxes; Other colors: predicted bounding boxes.}
	\label{fig_cat}
	\vspace{-10pt}
\end{figure*}

\subsection{Analysis}
Although the proposed DGQP module has been shown to improve the performance of dense object detectors, we would also like to understand how its mechanism operates. 

\noindent \textbf{DGQP Improves LQE}: To assess whether DGQP is able to benefit the estimation of localization quality, we first obtain the predicted IoUs (given by four representative models with IoU as the quality estimation targets) and their corresponding real IoUs over all the positive samples on COCO {\tt minival}. Then we calculate their Pearson Correlation Coefficient (PCC) in Table~\ref{tab_error}. It demonstrates that DGQP in GFLV2 indeed improves the linear correlation between the estimated IoUs and the ground-truth ones by a considerable margin (+0.26) against GFLV1, which eventually leads to an absolute 0.9 AP gain.

\begin{figure}[t]
	\vspace{-6pt}
	\begin{center}
		\setlength{\fboxrule}{0pt}
		\fbox{\includegraphics[width=0.44\textwidth]{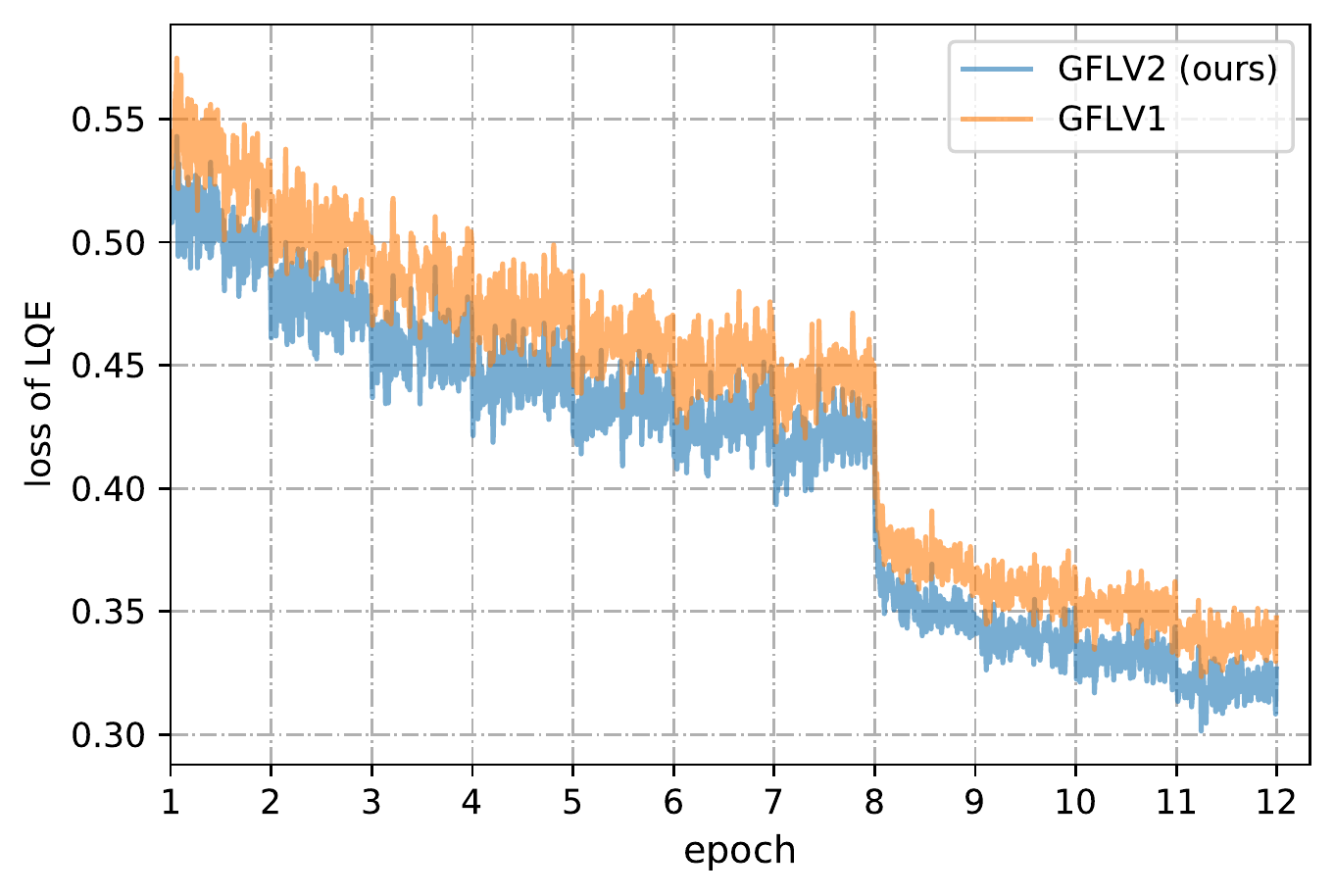}}
	\end{center}	
	\captionsetup{font={small}}
	\vspace{-24pt}
	\caption{Comparisons of losses on LQE between GFLV1 and GFLV2. DGQP helps to ease the learning difficulty with lower losses during training.}
	\label{fig_loss_iou}
	\vspace{-12pt}
\end{figure}

\noindent \textbf{DGQP Eases the Learning Difficulty}: Fig.~\ref{fig_loss_iou} provides the visualization of the training losses on LQE scores, where DGQP in GFLV2 successfully accelerates the training process and converges to lower losses.

\begin{figure}[t]
	\vspace{-6pt}
	\begin{center}
		\setlength{\fboxrule}{0pt}
		\fbox{\includegraphics[width=0.28\textwidth]{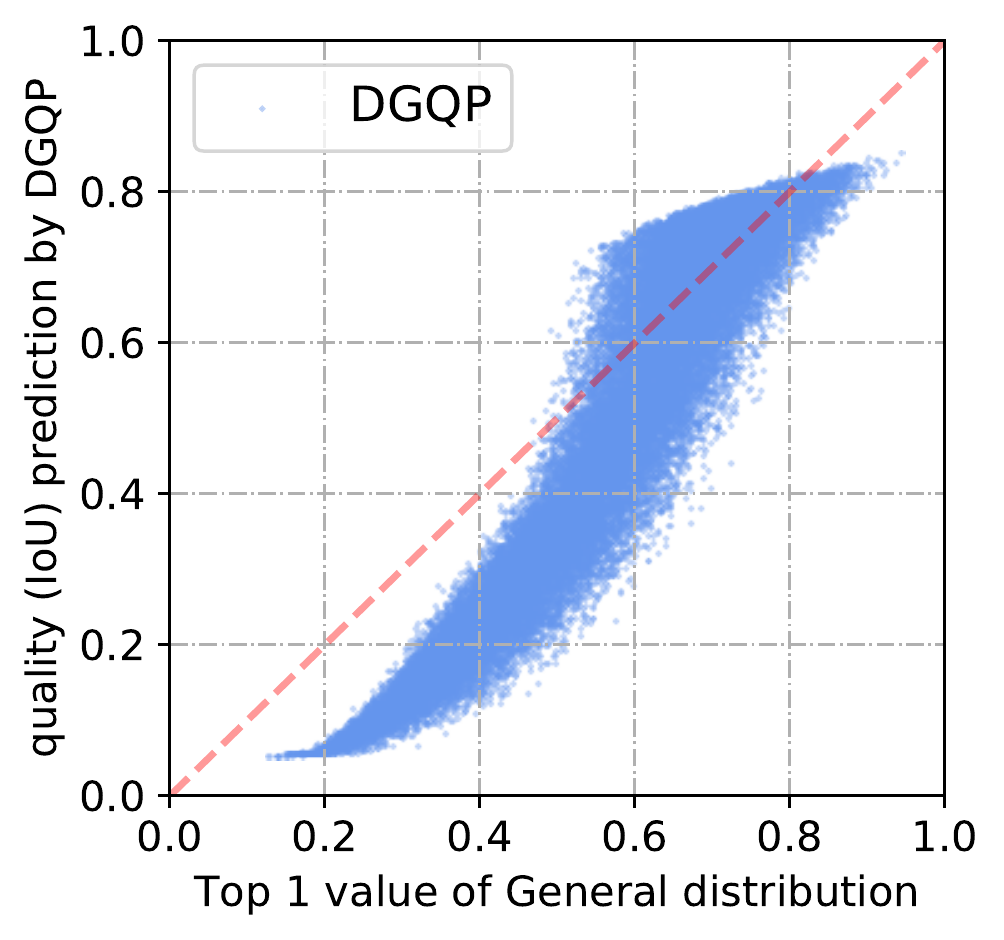}}
	\end{center}	
	\captionsetup{font={small}}
	\vspace{-22pt}
	\caption{Visualization of the input Top-1 values (mean of four sides) from the learned distribution and the output IoU quality prediction given by DGQP.}
	\label{fig_analysis_subnetwork}
	\vspace{-8pt}
\end{figure}

\noindent \textbf{Visualization on Inputs/Outputs of DGQP}: To study the behavior of DGQP, we plot its inputs and corresponding outputs in Fig.~\ref{fig_analysis_subnetwork}. For a better view, we select the mean Top-1 values to represent the input statistics. It is observed that the outputs are highly correlated with inputs as expected.

\noindent \textbf{Training/Inference Efficiency}: We also compare the training and inference efficiency among recent state-of-the-art dense detectors in Table~\ref{tab_efficiency}. Note that PAA \cite{kim2020probabilistic}, RepPointsV2 \cite{chen2020reppoints} and BorderDet \cite{qiu2020borderdet} bring an inevitable time overhead (52\%, 65\%, and 22\% respectively) during training, and the latter two also sacrifice inference speed by 30\% and 14\%, respectively. In contrast, our proposed GFLV2 can achieve top performance ($\sim$41 AP) while still maintaining the training and inference efficiency.


\begin{table}
	\small
	\centering
	\renewcommand\arraystretch{1.1}
	\newcommand{\tabincell}[2]{\begin{tabular}{@{}#1@{}}#2\end{tabular}}
	\resizebox{0.49\textwidth}{!}{
		\begin{tabular}{l|c|c|c}
			\hline
			Method & AP & Training Hours $\downarrow$ & Inference FPS $\uparrow$ \\
			\hline\hline
			ATSS$^*$ \cite{zhang2019bridging} & 39.9 &  \ \ 8.2 & 19.4  \\ 
			GFLV1    \cite{li2020generalized} & 40.2 &  \ \ 8.2 & {19.4}  \\ 
			PAA       \cite{kim2020probabilistic} & 40.4 & \ \ \ \ \ \ \ \ \ \ \  12.5 \textcolor{red}{\scriptsize (+52\%)} & {19.4}  \\
			RepPointsV2 \cite{chen2020reppoints} & 41.0 & \ \ \ \ \ \ \ \ \ \ \ 14.4  \textcolor{red}{\scriptsize (+65\%)} &  \ \ \ \ \ \ \ \ \ \    13.5 \textcolor{red}{\scriptsize (-30\%)}  \\ 
			BorderDet \cite{qiu2020borderdet} & 41.4 & \ \ \ \ \ \ \ \ \ \ \ 10.0 \textcolor{red}{\scriptsize (+22\%)} & \ \ \ \ \ \ \ \ \ \  16.7 \textcolor{red}{\scriptsize (-14\%)} \\ 
			\hline
			GFLV2 \textbf{(ours)} & 41.1 &  \ \ 8.2 & {19.4} \\ 
			\hline
		\end{tabular}
	}
\captionsetup{font={small}}
	\vspace{-8pt}
	\caption{Comparisons of training and inference efficiency based on ResNet-50 backbone. ``Training Hours'' is evaluated on 8 GeForce RTX 2080Ti GPUs under standard 1x schedule (12 epochs). $^*$ denotes the application of Classification-IoU Joint Representation.} 
	\vspace{-12pt}
	\label{tab_efficiency}
\end{table}

\noindent \textbf{Qualitative Results}: In Fig.~\ref{fig_cat}, we qualitatively demonstrate the mechanism how GFLV2 makes use of its more reliable IoU quality estimations to maintain accurate predictions during NMS. Unfortunately for other detectors, high-quality candidates are wrongly suppressed due to their relatively lower localization confidences, which eventually leads to a performance degradation.

\section{Conclusion}
In this paper, we propose to learn reliable localization quality estimation, through the guidance of statistics of bounding box distributions. It is an entirely new and completely different perspective in the literature, which is also conceptually effective as the information of distribution is highly correlated to the real localization quality. Based on it we develop a dense object detector, namely GFLV2. Extensive experiments and analyses on COCO dataset further validate its effectiveness, compatibility and efficiency. We hope GFLV2 can serve as simple yet effective baseline for the community.


{\small
\bibliographystyle{ieee_fullname}
\bibliography{egbib}
}

\end{document}